\documentclass{article}

\usepackage{microtype}
\usepackage{graphicx}
\usepackage{subcaption}
\usepackage{booktabs} 

\usepackage{hyperref}

\usepackage[preprint]{icml2026}

\usepackage{amsmath}
\usepackage{amssymb}
\usepackage{mathtools}
\usepackage{amsthm}

\usepackage[capitalize,noabbrev]{cleveref}

\theoremstyle{plain}

\theoremstyle{definition}

\theoremstyle{remark}

\usepackage[textsize=tiny]{todonotes}

\icmltitlerunning{StreamFlow: Accelerating Rectified Flow}

\begin{document}

\twocolumn[
  \icmltitle{StreamFlow: Theory, Algorithm, and Implementation \\for High-Efficiency Rectified Flow Generation}



  \icmlsetsymbol{equal}{*}
  
  \begin{icmlauthorlist}
    \icmlauthor{Sen Fang}{rutgers,equal}
    \icmlauthor{Hongbin Zhong}{gatech,equal}
    \icmlauthor{Yalin Feng}{ntu}
    \icmlauthor{Yanxin Zhang}{wisc}
    \icmlauthor{Dimitris N. Metaxas}{rutgers}
  \end{icmlauthorlist}
  
  \icmlaffiliation{rutgers}{Rutgers University, New Jersey, USA}
  \icmlaffiliation{gatech}{Georgia Institute of Technology, Atlanta, Georgia, USA}
  \icmlaffiliation{ntu}{Nanyang Technological University, Singapore}
  \icmlaffiliation{wisc}{University of Wisconsin-Madison, Wisconsin, USA}
  
  \icmlcorrespondingauthor{Sen Fang}{sen.fang@rutgers.edu}

  \icmlkeywords{Machine Learning, ICML}

  \vskip 0.3in
]



\printAffiliationsAndNotice{}  

\begin{abstract}

New technologies such as Rectified Flow and Flow Matching have significantly improved the performance of generative models in the past two years, especially in terms of control accuracy, generation quality, and generation efficiency. However, due to some differences in its theory, design, and existing diffusion models, the existing acceleration methods cannot be directly applied to the Rectified Flow model. In this article, we have comprehensively implemented an overall acceleration pipeline from the aspects of theory, design, and reasoning strategies. This pipeline uses new methods such as batch processing with a new velocity field, vectorization of heterogeneous time-step batch processing, and dynamic TensorRT compilation for the new methods to comprehensively accelerate related models based on flow models. Currently, the existing public methods usually achieve an acceleration of 18\%, while experiments have proved that our new method can accelerate the 512*512 image generation speed to up to 611\%, which is far beyond the current non-generalized acceleration methods. Project page at \href{https://world-snapshot.github.io/StreamFlow/}{https://world-snapshot.github.io/StreamFlow/}.

\end{abstract}
\section{Introduction}

\begin{figure}[t]
    \centering
    \includegraphics[width=.48\textwidth]{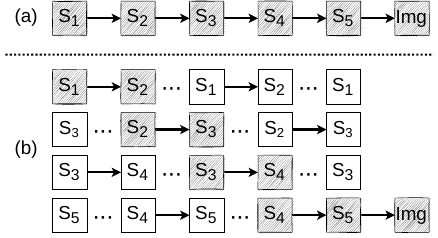}
    \captionof{figure}{\textbf{The differences between original denoising and batch denoising:} We can consider each line as a separate parallel operation queue. \textcolor{purple}{\textbf{(a)}} This is the original diffusion generation process. Suppose we need 5 steps to complete it, then denoising is sequential. \textcolor{purple}{\textbf{(b)}} Then some common acceleration methods involve using multiple parallel operation queues, with each queue only considering the requirements of a certain step. The {\color{gray}grey area} represents the complete process of a certain generation. 
    }
    \label{fig:cover}   
    \vspace{-12pt}
\end{figure}%

\textbf{Rectified Flow} \cite{liu2022flow} and \textbf{Flow Matching} \cite{lipman2022flow} are new generation methods that became popular about two years ago. Existing generation processes (such as \emph{diffusion} \cite{ho2020denoising, song2020denoising}) generally learn to gradually shift from a Gaussian noise distribution to the real distribution/image distribution. They usually require \emph{several dozen steps} to predict the next noise from the previous noise. However, these two new methods attempt to provide a \textit{velocity field}, allowing the model to directly predict the direction of the next distribution and the speed for each step, and then try to learn a more direct path between the distributions, enabling good results to be achieved in \textbf{4-step generation} \cite{yan2024perflow,liu2023instaflow}. 

Previously, the generation based on diffusion models has had many limitations in terms of theory and mechanism, resulting in relatively slow generation efficiency. With the popularity of diffusion models in areas such as augmented/virtual reality, style transfer, and real-time action generation, there have been works called \textbf{StreamDiffusion} \cite{kodaira2025streamdiffusion} that attempt to solve this problem. As shown in Fig. \ref{fig:cover}, they achieve this by creating a \textit{pipeline}, by decoupling noise addition and denoising stages and mapping them to parallel pipeline queues, which makes it particularly fast in accelerating the generation of diffusion models.

However, when attempting to transfer these diffusion-oriented acceleration frameworks to Rectified Flow models, we encounter fundamental mismatches at the level of representation, scheduling semantics, and execution structure: \textbf{(1)} As mentioned in the first paragraph, their theories are not consistent. The content they generate at each step and the process of generating also differ. For example, they deal with \textit{time fields} rather than noise. \textbf{(2)} In terms of scheduling strategies and other policies, they use the \textit{time step} to dynamically divide the progress of the content and then guide the generation process. \textbf{(3)} When we attempt to make them compatible with the existing compilation acceleration methods \cite{tensorrt2024}, due to their distinct structures and the fact that they change the structure depending on the settings, this often leads to \emph{compilation failures}.

To address these issues, we first, based on the generation theory of the Flow model, attempted to create a \emph{batch processing of velocity fields}. For different time steps as inputs, they were divided into different processing steps, conceptually treating them as a type of \textbf{operation step}. Secondly, we developed \textit{vectorized heterogeneous time-step batch processing} for these operation steps. According to the sequence of these different times, they were dynamically filled into the pipeline for processing. Finally, we regarded the entire framework as a dynamic and structurally changing generation process, and further accelerated it using \texttt{TensorRT} \cite{tensorrt2024} compilation.

However, when we actually implemented it, some new challenges emerged\footnote{These new challenges are specific to flow models and cannot be solved by any traditional acceleration framework.}:\textbf{(1)} Rectified Flow employs variable-length time steps, leading to misaligned and non-uniform processing stages.
 \textbf{(2)} The existing scheduler wrapper is unable to handle the \textit{vectorized output} and \textit{differentiable time steps}. \textbf{(3)} Even if \texttt{TensorRT} cannot simply achieve the functions we envisioned, the compilation is not only slow but also prone to crashing.

Therefore, we developed a batch processing system for the velocity field and made it compatible. Based on the theory of the flow model, we regarded the velocity field as \textit{random step-length noise steps}. Then we \textbf{decoupled the binding} of the noise steps with the pipeline, allowing the parallel queues to have any step-length velocities, and developed a scheduler that can handle such situations. For compilation-based acceleration, we added model construction compilation to automatically process structural information during compilation, and then our compilation script \textit{adaptively processes} this information for compilation.

Therefore, our contributions can be summarized as follows:
\begin{itemize}
    \item \textbf{Large-scale batch velocity field processing} with operational steps that \emph{decouple the binding} between step size and velocity field, featuring an adaptive pipeline with queue management for continuous velocity field output.
    \item A \textbf{scheduler that vectorizes heterogeneous timestep batches}, enabling unified processing of outputs with varying step sizes for collaborative velocity field generation and flow model architectures.
    \item \textbf{Dynamic singular structure \texttt{TensorRT} compilation} with \emph{plug-and-play compatibility} for flow model characteristics, further accelerating flow models within our acceleration framework.
\end{itemize}

\begin{figure*}[t]
  \centering
   \includegraphics[width=0.96\linewidth]{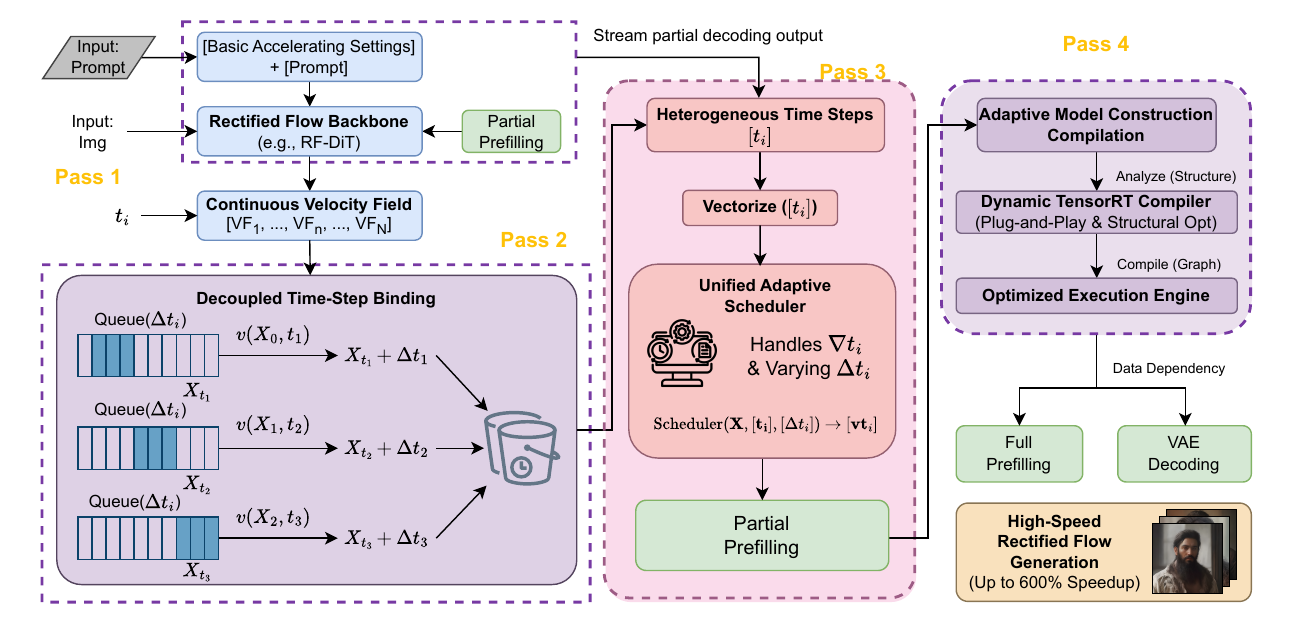}
   \vspace{-6pt}
   \caption{\textbf{Overview of our StreamFlow pipeline:} \textcolor{purple}{\textbf{(Pass 1)}} The process initiates with prompt processing via the Rectified Flow Backbone and partial prefilling to establish the continuous velocity field $v_t(X_t)$ based on the trajectory equation $dX_t = v_t(X_t)dt$. \textcolor{purple}{\textbf{(Pass 2)}} The Decoupled Time-Step Binding mechanism manages parallel queues with heterogeneous step sizes (e.g., $\Delta t_a, \Delta t_b$), independently applying velocity updates before converging via in-place aggregation for arbitrary batching. \textcolor{purple}{\textbf{(Pass 3)}} Heterogeneous time steps $[t_i]$ are vectorized and managed by the Unified Adaptive Scheduler, which handles varying temporal gradients and step sizes to synchronize velocity outputs for subsequent partial prefilling. \textcolor{purple}{\textbf{(Pass 4)}} We undergoes Adaptive Model Construction Compilation, utilizing a Dynamic TensorRT Compiler for plug-and-play compatibility and structural optimization to create an Optimized Execution Engine.}
   \label{fig:overview}
   \vspace{-4pt}
\end{figure*}

\section{Related Works}
\paragraph{Rectified Flow and Flow Matching:}
Flow Matching (FM) and Rectified Flow are recent paradigms that bridge diffusion models and continuous normalizing flows by directly regressing a velocity field $\mathbf{v}_\theta(\mathbf{x}_t, t)$ that maps Gaussian noise $\mathbf{x}_0 \sim \mathcal{N}(0, I)$ to data $\mathbf{x}_1 \sim p_{\text{data}}$ via the ODE $d\mathbf{x}_t = \mathbf{v}_\theta(\mathbf{x}_t, t) dt$~\cite{lipman2022flow, albergo2022building}. The key innovation lies in defining conditional probability paths that minimize transport cost. FM enables the use of optimal transport (straight-line) paths $\mathbf{x}_t = (1-t)\mathbf{x}_0 + t\mathbf{x}_1$, trained by minimizing the regression objective $\mathcal{L} = \mathbb{E}_{t, \mathbf{x}_0, \mathbf{x}_1} \| \mathbf{v}_\theta(\mathbf{x}_t, t) - (\mathbf{x}_1 - \mathbf{x}_0) \|^2$~\cite{lipman2022flow}. This formulation allows for fast inference through simple Euler integration $\mathbf{x}_1 \approx \mathbf{x}_0 + \mathbf{v}_\theta(\mathbf{x}_0, 0)$ in as few as one to four steps. Rectified Flow extends this by introducing an iterative ``reflow'' procedure, denoted as $\mathbf{x}_t^{(k+1)} = (1-t)\mathbf{z} + t \cdot \text{ODE-Solve}(\mathbf{z}, \mathbf{v}_\theta^{(k)})$, which progressively straightens the flow trajectories to reduce discretization error~\cite{liu2022flow}. Building on this, methods like InstaFlow successfully apply text-conditioned reflow to large-scale Stable Diffusion models, achieving high-quality one-step generation~\cite{liu2023instaflow}. These advancements lay the theoretical groundwork for our accelerated pipeline.

\vspace{-4pt}
\paragraph{Generative Image Models:}
Recent progress in generative modeling has been driven by diffusion, transformer, and flow-based architectures. DDPM~\cite{ho2020denoising} set the foundation for high-fidelity synthesis through iterative denoising, albeit with slow sampling. Latent Diffusion Models (LDM)~\cite{rombach2022high} reduce computation by operating in compressed latent space, while Diffusion Transformers (DiT)~\cite{peebles2023scalable} replace U-Nets with Vision Transformers to enhance scalability. Training these transformer-based models is further simplified by Representation Entanglement (REG)~\cite{wu2025representation}, which speeds up convergence by injecting semantic tokens from foundation models. Beyond classical diffusion, hybrid designs such as VA-VAE~\cite{yao2025reconstruction} employ pretrained vision models to regularize high-dimensional latents, alleviating the reconstruction–generation trade-off. For fast sampling, MeanFlow~\cite{geng2025mean} enables one-step generation by estimating average trajectory velocity, and PeRFlow~\cite{yan2024perflow} accelerates inference via piecewise linear flow segments, functioning as a plug-and-play module.  
Additionally, Visual Autoregressive Modeling (VAR)~\cite{tian2024visual} emerges as a scalable alternative with a next-scale prediction scheme. These advances motivate our new rectified flow acceleration framework.

\paragraph{Inference Acceleration:} To enable interactive generation \cite{Fang_2025_ICCV,fang2026signxcontinuoussignrecognition}, researchers have explored inference acceleration via step reduction optimizations. Multi-step diffusion models can be distilled to few-step or one-step generation~\cite{liu2022flow,liu2023instaflow}, while ODE-based solvers~\cite{liu2022flow,wu2025representation} reframe diffusion as neural ODEs for faster integration. Engineering optimizations like model quantization and TensorRT~\cite{tensorrt2024} can fuse layers and roughly double throughput by optimizing execution graphs. However, these assume fixed computation graphs and break under \emph{dynamic time steps} or changing batch sizes common in Flow methods.


We extend these ideas to rectified flow by designing fully vectorized, batched scheduling of continuous time steps and integrating dynamic TensorRT compilation to handle variable step sizes and scheduler incompatibilities that diffusion-specific accelerators do not face. Our unified pipeline boosts rectified flow generation by up to 611\%, far surpassing prior general-purpose methods. 

\section{Methodology}

  \subsection{Preliminaries and Problem Statement}

  \subsubsection{Rectified Flow and Velocity Fields}

  Rectified Flow \citep{liu2022flow,esser2024scalingrectifiedflowtransformers} formulates the generative process as learning a time-dependent velocity
  field $v_\theta(x_t, t)$ that transports samples from a noise distribution $\pi_0$ to the data distribution
  $\pi_1$ along straight paths. The flow is defined by the ODE $\frac{dx_t}{dt} = v_\theta(x_t, t)$ for $t \in
  [0, 1]$. The model learns to predict the velocity field by minimizing:

  \begin{equation}
  \mathcal{L} = \mathbb{E}_{x_0, x_1, t} \left[ \| v_\theta(x_t, t) - (x_1 - x_0) \|^2 \right]
  \label{eq:velocity_loss}
  \end{equation}

  where $x_t = t x_1 + (1-t) x_0$ represents the linear interpolation path. During inference, we discretize the
  continuous flow using $N$ steps with timesteps $\{t_i\}_{i=0}^{N-1}$: $x_{t_{i+1}} = x_{t_i} + \Delta t_i
  \cdot v_\theta(x_{t_i}, t_i)$ where $\Delta t_i = t_{i+1} - t_i$ is the step size.

  \subsubsection{Time-Windows Mechanism in PeRFlow}

  PeRFlow \citep{yan2024perflow} introduces a time-windows mechanism that divides the time interval $[0, 1]$
  into $K$ non-overlapping windows $\mathcal{W}_k = [t_k^s, t_k^e]$. For a given timestep $t_c \in
  \mathcal{W}_k$, the velocity is computed using window-specific parameters:

\begin{equation}
  \gamma_{s \rightarrow e} = \left( \frac{\alpha_{t_k^s}}{\alpha_{t_k^e}} \right)^{1/2}
\label{eq:gamma}
\end{equation}

\begin{equation}
  \lambda_t = \frac{\lambda_s (t_k^e - t_k^s)}{\lambda_s(t_c - t_k^s) + (t_k^e - t_c)}
\label{eq:lambda}
\end{equation}

\begin{equation}
  \eta_t = \frac{\eta_s (t_k^e - t_c)}{\lambda_s(t_c - t_k^s) + (t_k^e - t_c)}
\label{eq:time_windows}
\end{equation}

  where $\alpha_t$ represents the cumulative product of noise schedule coefficients, and $\lambda_s, \eta_s$ are
   window-dependent scaling factors derived from $\gamma_{s \rightarrow e}$. The predicted velocity is then
  $v_\theta(x_t, t) = (x_{pred}^{t_k^e} - x_t)/(t_k^e - t_c)$.

  \subsubsection{Challenges in Batch Acceleration}

  Prior work on accelerating diffusion models~\cite{lyu2022acceleratingdiffusionmodelsearly}, such as StreamDiffusion \citep{kodaira2025streamdiffusion},
  relies on batch denoising where all samples in a batch share the \emph{same timestep} $t$. This homogeneous
  batching allows for $\mathbf{X}_{t'} = f_\theta(\mathbf{X}_t, t)$ where $\mathbf{X}_t \in \mathbb{R}^{B \times
   C \times H \times W}$. However, this approach fails for Rectified Flow models due to three fundamental
  challenges:

  \textbf{Challenge 1: Velocity Field Batching.} Unlike noise prediction in diffusion models, velocity fields
  have varying step sizes $\Delta t_i$ that depend on the time-window structure. Standard schedulers cannot
  handle batched computations with different $\Delta t_i$ values.

  \textbf{Challenge 2: Heterogeneous Timesteps.} Pipeline-based acceleration requires processing samples at
  different denoising stages simultaneously, resulting in heterogeneous timestep batches $\mathbf{t} = [t_0,
  t_1, \ldots, t_{N-1}]$ where $t_i \neq t_j$ for $i \neq j$.

  \textbf{Challenge 3: Dynamic Compilation.} TensorRT and similar compilers assume static computational graphs.
  The time-window mechanism in Eq.~\ref{eq:time_windows} creates dynamic structures that violate this
  assumption, causing compilation failures.

  \subsection{StreamFlow: Architecture Overview}

  We propose StreamFlow, a comprehensive acceleration framework that addresses these challenges through three
  key innovations: (1) batched velocity field computation with vectorized time-windows, (2) heterogeneous
  timestep pipeline batching, and (3) dynamic TensorRT compilation. Figure~\ref{fig:overview} illustrates
  the overall pipeline.

  As shown in Fig. \ref{fig:overview}, the key insight is to \emph{decouple} the velocity field computation from step size constraints, enabling
  vectorized processing of heterogeneous timesteps while maintaining algorithmic correctness. This allows us to
  construct a pipeline where each UNet invocation processes $N$ samples at different denoising stages:

  \begin{equation}
  [\mathbf{x}_{t_0}, \mathbf{x}_{t_1}, \ldots, \mathbf{x}_{t_{N-1}}] \xrightarrow{UNet} [\mathbf{x}_{t_0}',
  \mathbf{x}_{t_1}', \ldots, \mathbf{x}_{t_{N-1}}']
  \label{eq:pipeline_batch}
  \end{equation}

  \subsection{Batched Velocity Field Computation}

  \subsubsection{Decoupling Step Size from Velocity Field}

  The vanilla PeRFlow \cite{yan2024perflow} scheduler processes timesteps sequentially, computing window parameters for each $t_i$
  individually. We reformulate this as a vectorized operation that processes a batch of timesteps $\mathbf{t} =
  [t_0, t_1, \ldots, t_{B-1}]$ simultaneously.

  First, we vectorize the window lookup operation. For each timestep $t_i$, we identify its corresponding window
   $\mathcal{W}_{k_i}$ through $k_i = \arg\max_k \{ t_i > t_k^e + \epsilon \}$ where $\epsilon$ is a numerical
  precision tolerance. This can be vectorized using masking operations: $\mathbf{M}_k = (\mathbf{t} > t_k^e +
  \epsilon)$ and $\mathbf{k} = \sum_{j=1}^K \mathbf{M}_j$.

  \subsubsection{Vectorized Window Parameter Calculation}

  Given the window assignments $\mathbf{k}$, we compute all window parameters in parallel. For a batch of $B$
  timesteps, we construct the window boundaries $\mathbf{t}^s = [t_{k_0}^s, \ldots, t_{k_{B-1}}^s]$ and
  $\mathbf{t}^e = [t_{k_0}^e, \ldots, t_{k_{B-1}}^e]$, then compute $\boldsymbol{\gamma} =
  (\boldsymbol{\alpha}(\mathbf{t}^s) / \boldsymbol{\alpha}(\mathbf{t}^e))^{1/2}$. The batch velocity field parameters are computed as:

\vspace{-28pt}
\begin{multline}
  \boldsymbol{\lambda}_s = \boldsymbol{\gamma}^{-1}, \quad
  \boldsymbol{\eta}_s = -\frac{(1 - \boldsymbol{\gamma}^2)^{1/2}}{\boldsymbol{\gamma}}, \\[-.3em]
  \text{denom} = \boldsymbol{\lambda}_s \odot (\mathbf{t} - \mathbf{t}^s) + (\mathbf{t}^e - \mathbf{t}), \\[-.3em]
  \boldsymbol{\lambda}_t = \frac{\boldsymbol{\lambda}_s \odot (\mathbf{t}^e - \mathbf{t}^s)}{\text{denom}}, \quad
  \boldsymbol{\eta}_t = \frac{\boldsymbol{\eta}_s \odot (\mathbf{t}^e - \mathbf{t})}{\text{denom}}
\label{eq:batch_velocity_params}
\end{multline}
\vspace{-20pt}

  where $\odot$ denotes element-wise multiplication.

\subsubsection{Batch Scheduler Design}

  We extend the standard scheduler with a \texttt{step\_batch} method that processes heterogeneous timesteps.
  For model outputs $\boldsymbol{\epsilon}_\theta \in \mathbb{R}^{B \times C \times H \times W}$ and latents
  $\mathbf{X} \in \mathbb{R}^{B \times C \times H \times W}$, we compute the predicted window endpoints
  $\mathbf{X}_{pred}^e = \boldsymbol{\lambda}_t \odot \mathbf{X} + \boldsymbol{\eta}_t \odot
  \boldsymbol{\epsilon}_\theta$, the velocity field $\mathbf{v}_\theta = (\mathbf{X}_{pred}^e -
  \mathbf{X})/(\mathbf{t}^e - \mathbf{t})$, and the next latents $\mathbf{X}_{next} = \mathbf{X} + \Delta
  \mathbf{t} \odot \mathbf{v}_\theta$ where $\Delta \mathbf{t} = \mathbf{t}_{next} - \mathbf{t}$.

  Algorithm~\ref{alg:batch_velocity} presents the complete batched velocity field computation.

  \begin{algorithm}[t]
  \caption{Batched Velocity Field Step}
  \label{alg:batch_velocity}
  \begin{algorithmic}[1]
  \REQUIRE Model output $\boldsymbol{\epsilon}_\theta$, latents $\mathbf{X}$, timesteps $\mathbf{t}$
  \ENSURE Updated latents $\mathbf{X}_{next}$
  \STATE $\mathbf{t}_c \gets \mathbf{t} / T_{max}$ \COMMENT{Normalize to $[0,1]$}
  \STATE Compute window assignments: $\mathbf{t}^s, \mathbf{t}^e \gets \text{WindowLookup}(\mathbf{t}_c)$
  \STATE $\boldsymbol{\gamma} \gets (\text{alphas\_cumprod}[\mathbf{t}^s] /
  \text{alphas\_cumprod}[\mathbf{t}^e])^{1/2}$
  \STATE $\boldsymbol{\lambda}_s \gets 1/\boldsymbol{\gamma}$, $\boldsymbol{\eta}_s \gets
  -(1-\boldsymbol{\gamma}^2)^{1/2}/\boldsymbol{\gamma}$
  \STATE $\text{denom} \gets \boldsymbol{\lambda}_s \odot (\mathbf{t}_c - \mathbf{t}^s) + (\mathbf{t}^e -
  \mathbf{t}_c)$
  \STATE $\boldsymbol{\lambda}_t \gets \boldsymbol{\lambda}_s \odot (\mathbf{t}^e - \mathbf{t}^s) /
  \text{denom}$
  \STATE $\boldsymbol{\eta}_t \gets \boldsymbol{\eta}_s \odot (\mathbf{t}^e - \mathbf{t}_c) / \text{denom}$
  \STATE $\mathbf{X}_{pred}^e \gets \boldsymbol{\lambda}_t \odot \mathbf{X} + \boldsymbol{\eta}_t \odot
  \boldsymbol{\epsilon}_\theta$
  \STATE $\mathbf{v}_\theta \gets (\mathbf{X}_{pred}^e - \mathbf{X}) / (\mathbf{t}^e - \mathbf{t}_c)$
  \STATE $\mathbf{t}_{next} \gets \text{GetNextTimesteps}(\mathbf{t})$
  \STATE $\Delta \mathbf{t} \gets (\mathbf{t}_{next} - \mathbf{t}) / T_{max}$
  \STATE $\mathbf{X}_{next} \gets \mathbf{X} + \Delta \mathbf{t} \odot \mathbf{v}_\theta$
  \STATE \textbf{return} $\mathbf{X}_{next}$
  \end{algorithmic}
  \end{algorithm}

  \textbf{Complexity Analysis.} The vanilla scheduler has time complexity $O(N \cdot T)$ for $N$ images with $T$
   denoising steps each. Our batched scheduler reduces this to $O((N+T) \cdot K)$ where $K$ is the number of
  time windows (typically $K=4$), achieving significant speedup when $N \gg K$.

  \subsection{Heterogeneous Timestep Pipeline Batching}

  \subsubsection{Pipeline Architecture and Buffer Management}

  We construct a pipeline that processes $N$ samples at different denoising stages concurrently. Let $S_i$
  denote the $i$-th denoising stage with timestep $t_i$. We maintain a latent buffer $\mathcal{B} =
  \{z_1^{(S_1)}, z_2^{(S_2)}, \ldots, z_{N-1}^{(S_{N-1})}\}$ storing intermediate latents at different stages.

  At each iteration $j$, a new sample $z_{new}$ enters the pipeline at stage $S_0$. The pipeline batch is
  constructed as $\mathcal{P}_j = [z_{new}^{(S_0)}, z_1^{(S_1)}, \ldots, z_{N-1}^{(S_{N-1})}]$ with
  corresponding heterogeneous timesteps $\mathbf{t}_j = [t_0, t_1, \ldots, t_{N-1}]$. A single UNet forward pass
   processes all stages: $\boldsymbol{\epsilon}_\theta(\mathcal{P}_j, \mathbf{t}_j, c) = [\epsilon_0,
  \epsilon_1, \ldots, \epsilon_{N-1}]$ where $c$ is the conditioning.

  \subsubsection{Asynchronous Queue Processing}

  After the UNet forward pass, we apply the batched scheduler (Algorithm~\ref{alg:batch_velocity}) to obtain
  updated latents $\mathcal{P}_j'$. The pipeline buffer is then updated via a shift operation:
  $\mathcal{B}_{j+1} = [z_{new}'^{(S_1)}, z_1'^{(S_2)}, \ldots, z_{N-2}'^{(S_{N-1})}]$. The completed sample
  $z_{N-1}'^{(S_N)}$ exits the pipeline for VAE~\cite{kingma2022autoencodingvariationalbayes} decoding. This asynchronous processing ensures continuous
  throughput: once the pipeline is full (after $N-1$ warmup iterations), each UNet call produces one complete
  sample.

  \subsubsection{Throughput Analysis}

  Let $C_{UNet}$ denote the cost of a single UNet forward pass, $C_{VAE}$ the VAE decoding cost, and $C_{sched}$
   the scheduler cost. For generating $M$ images, the vanilla approach requires $T_{vanilla} = M \cdot (N \cdot
  C_{UNet} + N \cdot C_{sched} + C_{VAE})$, while StreamFlow requires:

\begin{equation}
  \begin{aligned}
  T_{ours} = &(M + N - 1) \cdot C_{UNet} + (M + N - 1) \cdot C_{sched} \\
  &+ M \cdot C_{VAE} \approx M \cdot (C_{UNet} + C_{sched} + C_{VAE})
  \end{aligned}
  \label{eq:time_ours}
\end{equation}

  for $M \gg N$. The speedup factor is approximately $N$ when $C_{UNet}$ dominates. For $N=4$ denoising steps,
  we achieve theoretical $4\times$ speedup. Algorithm~\ref{alg:pipeline_batch} describes the complete pipeline
  batching procedure.

  \begin{algorithm}[t]
  \caption{Pipeline Batch Denoising}
  \label{alg:pipeline_batch}
  \begin{algorithmic}[1]
  \REQUIRE Conditioning $c$, number of images $M$, denoising steps $N$
  \ENSURE Generated images $\{I_1, \ldots, I_M\}$
  \STATE Initialize: $\mathcal{B} \gets \emptyset$, $\mathbf{t} \gets [t_0, \ldots, t_{N-1}]$,
  $\text{initialized} \gets \text{False}$
  \FOR{$j = 1$ to $M + N - 1$}
      \STATE $z_{new} \sim \mathcal{N}(0, I)$; $\mathcal{P}_j \gets [z_{new}, \mathcal{B}]$
      \IF{$|\mathcal{P}_j| = N$ and not initialized}
          \STATE $\text{initialized} \gets \text{True}$
      \ENDIF
      \STATE $\mathbf{t}_{batch} \gets \mathbf{t}[0:|\mathcal{P}_j|]$
      \STATE Apply CFG if needed: $\mathcal{P}_{in}, \mathbf{t}_{in}, c_{in} \gets
  \text{ApplyCFG}(\mathcal{P}_j, \mathbf{t}_{batch}, c)$
      \STATE $\boldsymbol{\epsilon}_\theta \gets \text{UNet}(\mathcal{P}_{in}, \mathbf{t}_{in}, c_{in})$
      \STATE $\boldsymbol{\epsilon}_\theta \gets \text{HandleCFG}(\boldsymbol{\epsilon}_\theta)$
      \STATE $\mathcal{P}_j' \gets \text{BatchVelocityStep}(\boldsymbol{\epsilon}_\theta, \mathcal{P}_j,
  \mathbf{t}_{batch})$
      \IF{initialized}
          \STATE $I_{j-N+1} \gets \text{VAE.decode}(\mathcal{P}_j'[-1])$; $\mathcal{B} \gets
  \mathcal{P}_j'[0:-1]$
      \ELSE
          \STATE $\mathcal{B} \gets \mathcal{P}_j'$
      \ENDIF
  \ENDFOR
  \STATE \textbf{return} $\{I_1, \ldots, I_M\}$
  \end{algorithmic}
  \end{algorithm}
  \vspace{-12pt}

\begin{figure*}[t]
  \centering
   \includegraphics[width=0.96\linewidth]{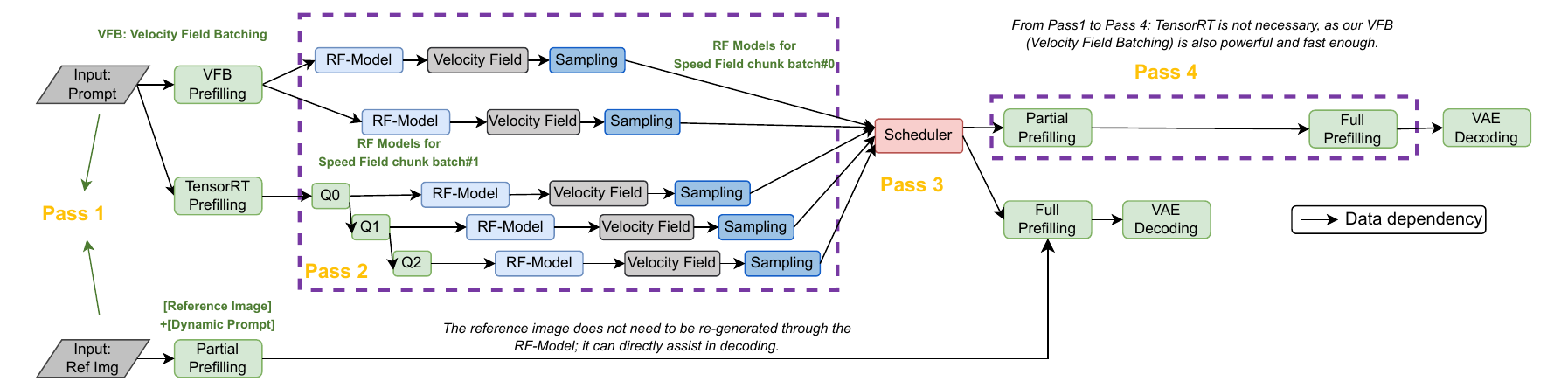}
   \vspace{-6pt}
   \caption{\textbf{Detailed implementation of StreamFlow's batched velocity field processing and heterogeneous timestep pipeline.} Starting from prompt and reference image inputs, \textcolor{purple}{\textbf{Pass 1}} performs TensorRT-optimized/VFB (Velocity Field Batching) prefilling to establish initial latent representations. \textcolor{purple}{\textbf{Pass 2}} demonstrates the core velocity field batch processing: multiple RF model instances process parallel queues (Q0, Q1, Q2) with heterogeneous timesteps, where each queue independently computes velocity fields before synchronization through reranking modules. \textcolor{purple}{\textbf{Pass 3}} shows the asynchronous queue management: the unified adaptive scheduler coordinates multiple parallel streams, each progressing through partial prefilling → full prefilling → VAE decoding stages at different rates. \textcolor{purple}{\textbf{Pass 4}} illustrates the dynamic dependency graph that enables proper data flow across heterogeneous processing stages. 
   Data dependency arrows show how velocity field outputs from Pass 2 feed into the scheduler, which then manages the asynchronous progression of multiple generation streams in Pass 3, achieving continuous throughput without blocking.}
   \label{fig:method}
   \vspace{-4pt}
\end{figure*}

  \subsection{Dynamic TensorRT Compilation}

  \subsubsection{Challenge: Static Compilation vs. Dynamic Structure}

TensorRT \citep{tensorrt2024} is a deep learning inference optimizer that compiles neural networks into optimized execution graphs. However, it assumes \emph{static} inputs~\cite{chen2018tvmautomatedendtoendoptimizing}: all samples in a batch must have identical shapes and processing paths. Our heterogeneous timestep batching violates this assumption because different timesteps may trigger different execution branches~\cite{shen2021nimbleefficientlycompilingdynamic} in the time-windows mechanism.

  As shown in Fig. \ref{fig:method}, formally, let $G_t$ denote the computational graph for timestep $t$. For a batch with heterogeneous timesteps
  $\mathbf{t} = [t_0, \ldots, t_{N-1}]$, the combined graph is $G_{batch} = \bigcup_{i=0}^{N-1} G_{t_i}$. If
  $G_{t_i} \neq G_{t_j}$ for some $i, j$, TensorRT compilation fails because it cannot determine a single static
   graph at compile time.

  \subsubsection{Runtime Adaptive Strategy}

  We address this by introducing a runtime detection and decomposition mechanism. At each forward pass, we check
   whether the timestep batch is homogeneous: $\text{is\_homogeneous}(\mathbf{t}) = \mathbb{I}[t_i = t_j \;
  \forall i, j]$.

  \textbf{Case 1: Homogeneous batch.} All timesteps are identical. We directly invoke the compiled TensorRT
  engine: $\boldsymbol{\epsilon}_\theta = \text{TensorRT}(\mathcal{P}, t, c)$. This is the common case during
  warmup and when using batch generation without pipeline.

  \textbf{Case 2: Heterogeneous batch.} Timesteps differ. We decompose the batch into individual samples:
  $\boldsymbol{\epsilon}_\theta = \bigoplus_{i=0}^{N-1} \text{TensorRT}(z_i, t_i, c_i)$ where $\bigoplus$
  denotes concatenation along the batch dimension. While this introduces $N$ TensorRT calls instead of 1, the
  per-call overhead is minimal ($<$1ms), and the overall speedup from TensorRT optimization (typically
  $2-3\times$) still provides substantial gains.

  \textbf{Performance trade-off.} Let $O_{TRT}$ be the TensorRT overhead per call. The total cost becomes
  $C_{heterogeneous} = N \cdot (O_{TRT} + C_{UNet}^{TRT}) \ll N \cdot C_{UNet}^{vanilla}$. Since $C_{UNet}^{TRT}
   \approx 0.3 \cdot C_{UNet}^{vanilla}$ and $O_{TRT} \approx 0.001 \cdot C_{UNet}^{vanilla}$, we still achieve
  $\approx 3\times$ speedup despite the decomposition.

  Algorithm~\ref{alg:tensorrt} presents the complete adaptive TensorRT forward pass.

\textbf{Plug-and-play compatibility.} Our approach wraps the TensorRT engine without modifying the compilation process itself. This ensures compatibility with any TensorRT-compiled UNet, requiring zero changes to existing acceleration pipelines. The runtime overhead of homogeneity checking is negligible ($<$0.01ms).

\begin{algorithm}[t]
\caption{TensorRT Compatible Forward}
\label{alg:tensorrt}
\begin{algorithmic}[1]
\REQUIRE Latents $\mathcal{P}$, timesteps $\mathbf{t}$, conditioning $c$
\ENSURE Model output $\boldsymbol{\epsilon}_\theta$
\STATE $\mathbf{t}_{\text{unique}} \gets \text{Unique}(\mathbf{t})$
\IF{$|\mathbf{t}_{\text{unique}}| = 1$}
    \STATE $\boldsymbol{\epsilon}_\theta \gets \text{TensorRT\_Forward}(\mathcal{P}, \mathbf{t}[0], c)$ \COMMENT{Homogeneous case}
\ELSE
    \STATE $\text{results} \gets [\text{TensorRT\_Forward}(\mathcal{P}[i], \mathbf{t}[i], c[i]) \text{ for } i \in [0, B-1]]$ \COMMENT{Heterogeneous decomposition}
    \IF{$\text{isinstance}(\text{results}[0], \text{tuple})$}
        \STATE $\boldsymbol{\epsilon}_\theta \gets \text{ConcatenateTuples}(\text{results})$
    \ELSE
        \STATE $\boldsymbol{\epsilon}_\theta \gets \text{Concatenate}(\text{results}, \text{dim}=0)$
    \ENDIF
\ENDIF
\STATE \textbf{return} $\boldsymbol{\epsilon}_\theta$
\end{algorithmic}
\end{algorithm}

\section{Experiments}

\begin{figure*}[t]
    \centering
    \includegraphics[width=0.95\textwidth]{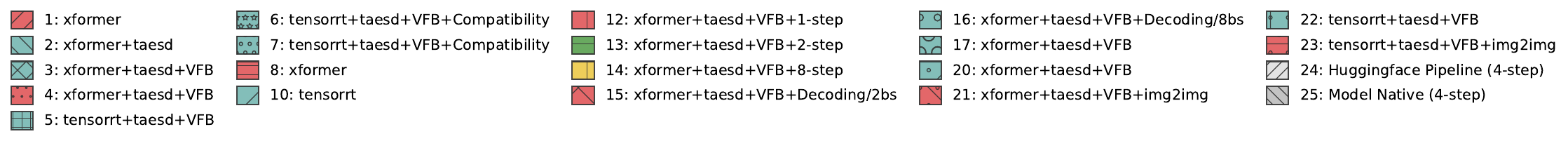}
    \begin{subfigure}[b]{0.15\textwidth}\centering
      \includegraphics[width=\linewidth]{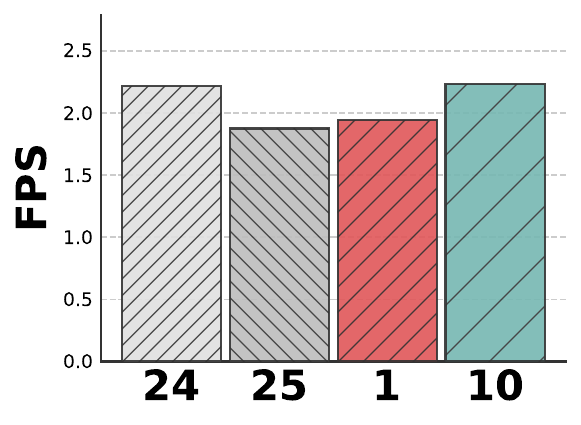}\caption{\tiny Just by switching to our pipeline, a slight improvement can be observed; enabling VFB can lead to a more significant enhancement.}
      \label{fig:abl:a}
    \end{subfigure}\hfill
    \begin{subfigure}[b]{0.15\textwidth}\centering
      \includegraphics[width=\linewidth]{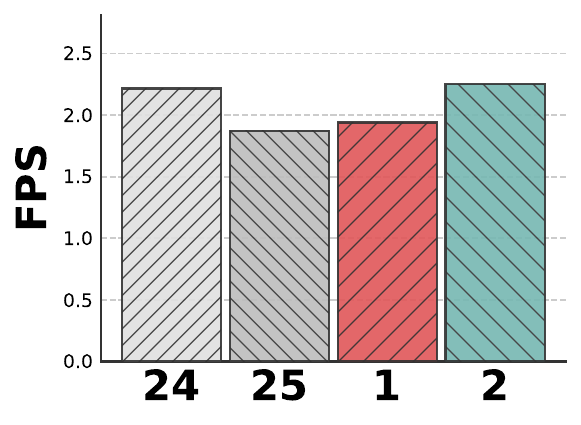}\caption{\tiny If we replace VFB with our pipeline's taesd (A tiny VAE distilled from SD VAE), similar improvements with (a) can also be observed.}
      \label{fig:abl:b}
    \end{subfigure}\hfill
    \begin{subfigure}[b]{0.15\textwidth}\centering
      \includegraphics[width=\linewidth]{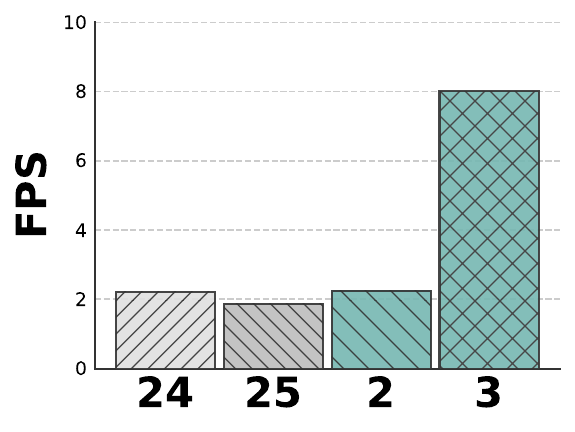}\caption{\tiny If VFB is enabled based on the (b), a significant improvement can be clearly observed, as the decoder was previously the weak point.}
      \label{fig:abl:c}
    \end{subfigure}\hfill
    \begin{subfigure}[b]{0.15\textwidth}\centering
      \includegraphics[width=\linewidth]{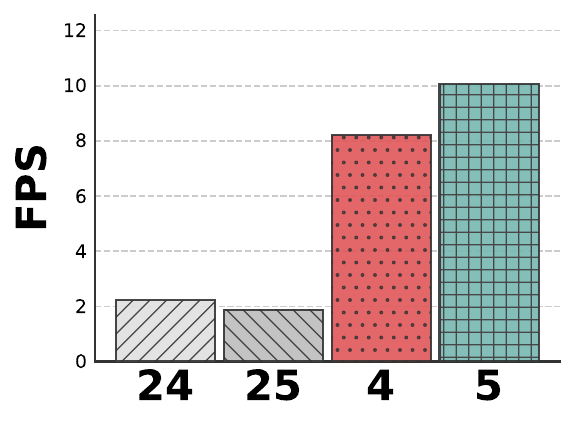}\caption{\tiny Based on (c), we attempted to start TensorRT. We could observe that the optimization of denoising led to a further improvement.}
      \label{fig:abl:d}
    \end{subfigure}\hfill
    \begin{subfigure}[b]{0.15\textwidth}\centering
      \includegraphics[width=\linewidth]{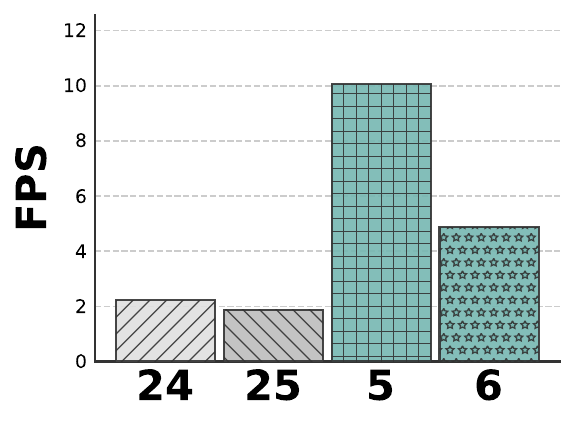}\caption{\tiny The compatibility layer has slowed down the speed improvement, but it has enhanced the stability and adaptability of the compilation.}
      \label{fig:abl:e}
    \end{subfigure}\hfill
    \begin{subfigure}[b]{0.15\textwidth}\centering
      \includegraphics[width=\linewidth]{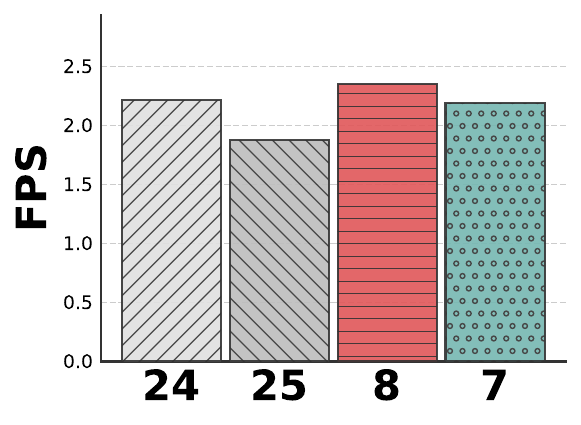}\caption{\tiny Just by comparing the different architectures of our pipelines, we can see that our pipelines are always higher than the baseline in any case.}
      \label{fig:abl:f}
    \end{subfigure}
    \begin{subfigure}[b]{0.15\textwidth}\centering
      \includegraphics[width=\linewidth]{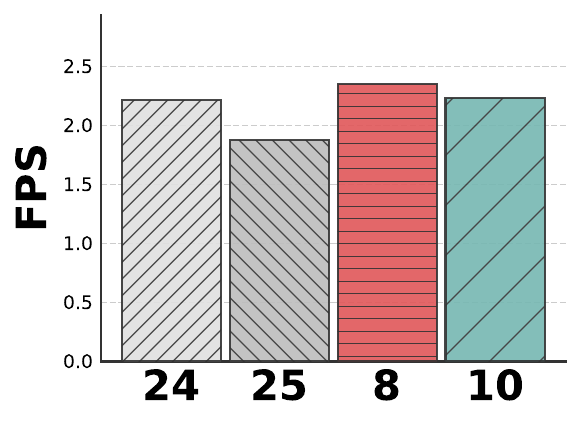}\caption{\tiny We enabled VFB in our pipeline under the same xformer architecture, and found that even though the architecture was changed, our VFB still accelerated.}
      \label{fig:abl:g}
    \end{subfigure}\hfill
    \begin{subfigure}[b]{0.15\textwidth}\centering
      \includegraphics[width=\linewidth]{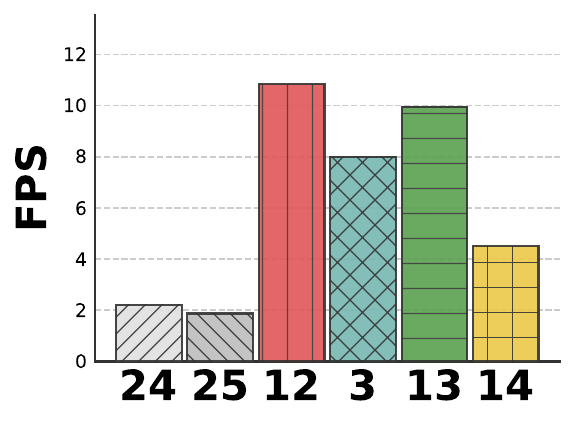}\caption{\tiny As the number of reasoning steps increases, our speed gradually decreases. However, after we accelerate, even with 8 steps, it is still faster than the basic 4 steps.}
      \label{fig:abl:h}
    \end{subfigure}\hfill
    \begin{subfigure}[b]{0.15\textwidth}\centering
      \includegraphics[width=\linewidth]{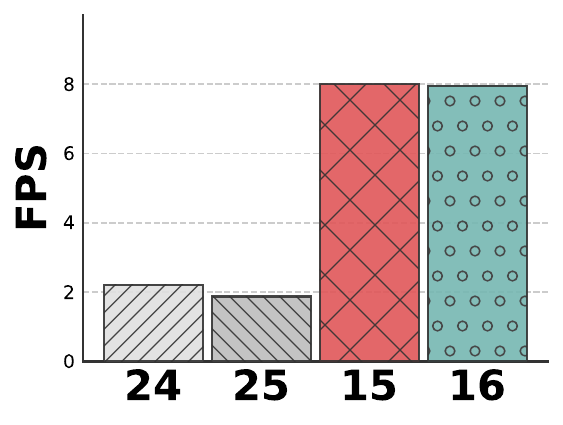}\caption{\tiny There is almost no difference, indicating that our generation potential variables are evolving very quickly, to the extent that the bottleneck actually lies in the micro-decoder.}
      \label{fig:abl:i}
    \end{subfigure}\hfill
    \begin{subfigure}[b]{0.15\textwidth}\centering
      \includegraphics[width=\linewidth]{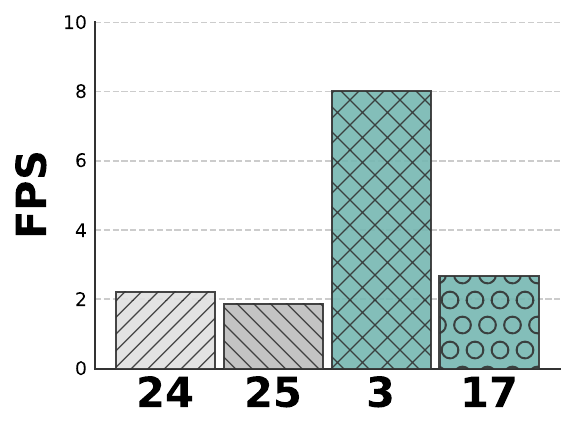}\caption{\tiny It shows that the changes made to StremDiffusion have been successfully inherited. The cfg file is not the bottleneck; instead, the native cfg file actually imposes a burden.}
      \label{fig:abl:j}
    \end{subfigure}\hfill
    \begin{subfigure}[b]{0.15\textwidth}\centering
      \includegraphics[width=\linewidth]{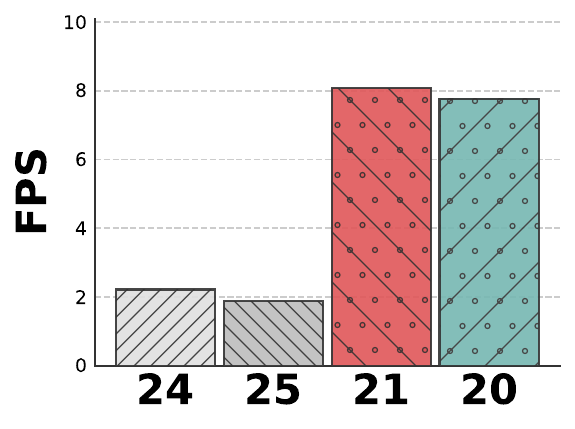}\caption{\tiny It can be seen that the full optimization of our code (in the xformer mode) is also beneficial for the img2img mode, and the performance is almost the same.}
      \label{fig:abl:k}
    \end{subfigure}\hfill
    \begin{subfigure}[b]{0.15\textwidth}\centering
      \includegraphics[width=\linewidth]{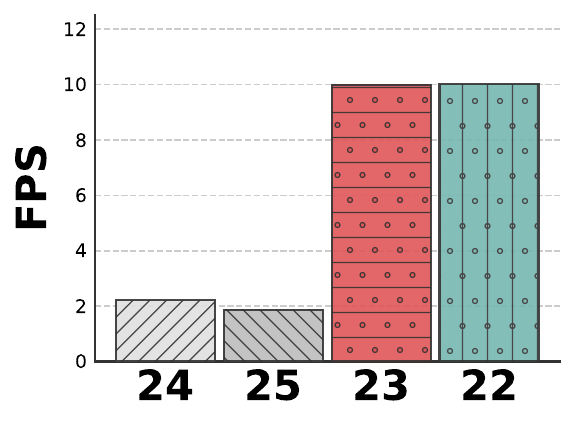}\caption{\tiny It can be seen that the full optimization of our code (in the tensorrt mode) is also beneficial for the img2img mode, and the performance is almost the same.}
      \label{fig:abl:l}
    \end{subfigure}

    \caption{\textbf{Ablation Study:} We conducted a detailed ablation study on all the components, and the experiments proved that all the components have a significantly higher average generation speed compared to the original model or the official pipeline of Huggingface. Additionally, in the fastest case, our peak improvement can reach 11/1.8, which is approximately a 611\% increase from the original speed.}
    \label{fig:ablations}  
    \vspace{-4pt}
\end{figure*}

Our work differs from previous acceleration frameworks which focus on optimizing different models\cite{ma2023deepcacheacceleratingdiffusionmodels,li2024distrifusiondistributedparallelinference}. Therefore, they do not compete with each other. Instead, our approach is compatible with the existing Rectified Flow models' acceleration methods\footnote{Accelerated methods for the traditional Rectified Flow model are generally non-transferable, require architecture changes, re-training or distillation.}\cite{liu2024instaflowstephighqualitydiffusionbased,yan2024perflowpiecewiserectifiedflow}. They can be used in combination rather than being mutually exclusive. Therefore, we mainly focused on the effectiveness and reliability of our method, and conducted detailed ablation evaluations, power and memory efficiency evaluations, scalability robustness evaluations, and quality evaluations, etc. We test our pipeline on a workstation equipped with 8x NVIDIA Quadro RTX 8000 GPUs (46GB VRAM each), Intel Xeon Silver 4116 CPU @ 2.10GHz, and Ubuntu 24.04 LTS for image generation. We measured the inference time for an average of 100 images, following the convention of previous studies \cite{kodaira2025streamdiffusion,feng2025streamdiffusionv2streamingdynamicinteractive}.

\begin{table}[t]
\centering
\caption{\textbf{Power/Memory Comparison:} Higher FPS is better; lower power and peak memory are better. The img2img mode here is a noise-free mode because there is a reference image.}
\label{tab:power_mem}
\resizebox{\linewidth}{!}{%
\begin{tabular}{lccc}
\toprule
Method & FPS $\uparrow$ & Power (W) $\downarrow$ & Peak Mem (MB) $\downarrow$ \\
\midrule
xformer+taesd+VFB+1-step             & 10.83 & 171.95 & 2976 \\
tensorrt+taesd+VFB       & 10.05 & 110.51 & 2666 \\
tensorrt+taesd+VFB+img2img           & 10.01 & 104.50 & 2694 \\
xformer+taesd+VFB   &  8.00 & 153.17 & 3012 \\
xformer+taesd+VFB+img2img            &  7.77 &  31.82 & 2966 \\
\midrule
Baseline HF \cite{von-platen-etal-2022-diffusers}              &  2.22 & 196.68 & 3696 \\
Baseline Vanilla \cite{yan2024perflow}         &  1.87 & 205.28 & 3788 \\
\bottomrule
\end{tabular}
}
\vspace{-4pt}
\end{table}

\subsection{Ablation Assessment}

Our framework inherits two architectural-level optimizations from Xformer\cite{zhang2024xformerhybridxshapedtransformer} and TensorRT \cite{tensorrt2024}. These can be switched, so there are two scenarios that need to be compared. Without any special indication, all are four-step inference speeds. VFB refers to Velocity Field Batching; taesd refers to the Tiny VAE \cite{kingma2022autoencodingvariationalbayes} included in the acceleration framework; By default, all images are set to a size of 512*512.

As demonstrated in Figure \ref{fig:ablations}, we conducted comprehensive ablation studies to validate each component's contribution. The baseline performance shows that the vanilla implementation achieves 1.87 FPS while the Hugging Face pipeline reaches 2.22 FPS. When we incrementally add optimizations starting with Xformer architecture (Fig \ref{fig:abl:a}) and taesd decoder (Fig \ref{fig:abl:b}), we observe modest improvements to 2.3 FPS, confirming that tiny VAE decoding represents a significant bottleneck\cite{rombach2022highresolutionimagesynthesislatent}. The most substantial gains emerge when enabling VFB with taesd (Fig \ref{fig:abl:c}), achieving 8.0 FPS—a 360\% improvement that validates our core insight that batched velocity field~\cite{lipman2023flowmatchinggenerativemodeling} processing is essential for Rectified Flow models. Further enabling TensorRT compilation (Fig \ref{fig:abl:d}) pushes performance to 10.05 FPS, while adding the compatibility layer (Fig \ref{fig:abl:e}) introduces a minor trade-off (9.8 FPS) but enhances stability across diverse architectures. Our experiments reveal important scalability insights: even with 8 inference steps, our optimized pipeline (4.5 FPS, Fig \ref{fig:abl:h}) outperforms the 4-step baseline (2.22 FPS), and the decoder batch size comparison (Fig \ref{fig:abl:i}) shows that latent generation is no longer the bottleneck—the system is now limited by VAE decoding throughput.

\begin{figure}[t]
    \centering
    \begin{subfigure}[b]{0.22\textwidth}\centering
      \includegraphics[width=\linewidth]{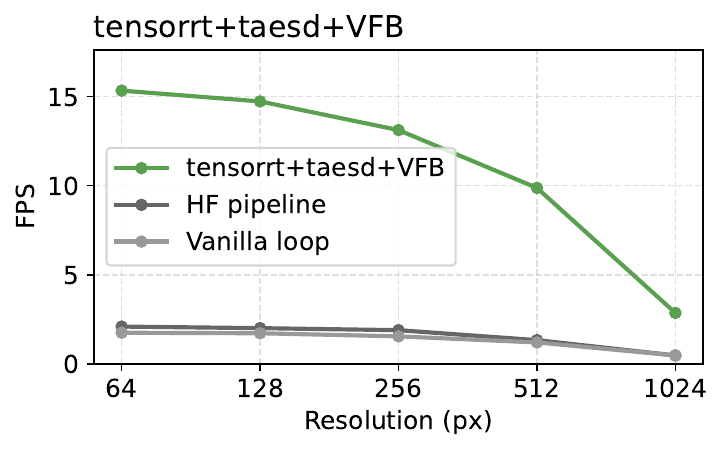}\vspace{1pt}\caption{\tiny Our method can still function after being compiled by TensorRT. However, it will be affected when the image size increases, but the baseline is more significantly impacted.}
      \label{fig:scable:a}
    \end{subfigure}\hfill
    \begin{subfigure}[b]{0.22\textwidth}\centering
      \includegraphics[width=\linewidth]{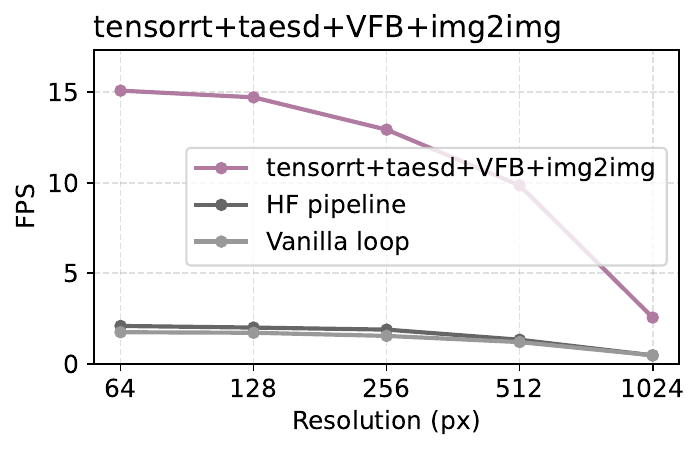}\caption{\tiny When in the img2img mode, we can observe that this has very little impact, which is also consistent with the performance of the ablation study.}
      \label{fig:scable:b}
    \end{subfigure}
    \begin{subfigure}[b]{0.22\textwidth}\centering
      \includegraphics[width=\linewidth]{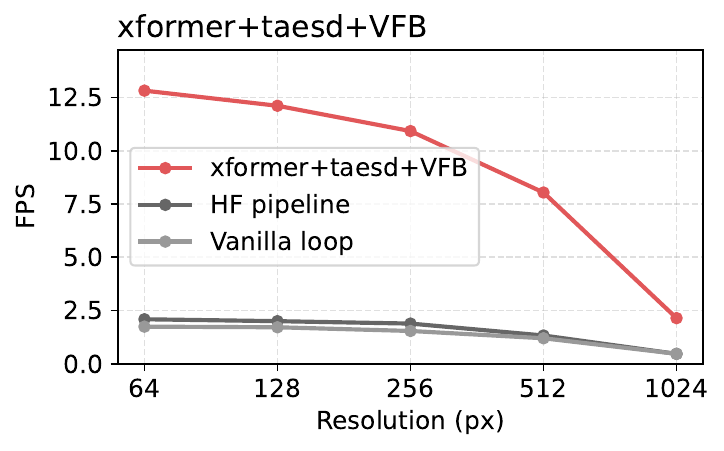}\caption{\tiny Under the xformer framework, our initial acceleration ratio is slightly lower than that of TensorRT, but the reduction in our ratio is slightly less than TensorRT.}
      \label{fig:scable:c}
    \end{subfigure}\hfill
    \begin{subfigure}[b]{0.22\textwidth}\centering
      \includegraphics[width=\linewidth]{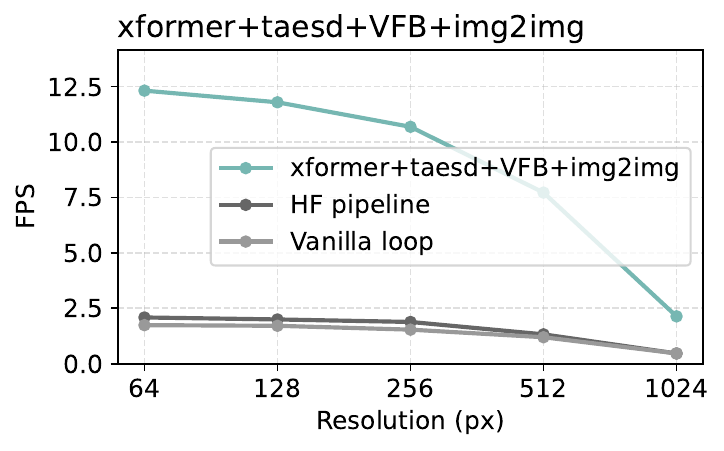}\caption{\tiny Then we can observe that the accelerated framework in the xformer mode performs quite robustly. Even if the baseline is significantly affected, we do not suffer much loss.}
      \label{fig:scable:d}
    \end{subfigure}
    \caption{\textbf{Scalability Study:} By changing the target size of the generated images, we found that our method has much stronger robustness compared to the previous methods. We are hardly affected by too much size variation. Even at relatively large sizes, we still achieve an acceleration of almost four to five times compared to the baseline. This not only demonstrates the robustness of our method, but also proves that our approach is a true method-to-strategy acceleration, rather than taking any expedient path.}
    \label{fig:scable}  
    \vspace{-12pt}
\end{figure}

\subsection{Efficiency Assessment}

Beyond raw speed improvements, we evaluate the resource efficiency of StreamFlow across power consumption and memory footprint metrics. Table \ref{tab:power_mem} presents comprehensive measurements for our key configurations alongside baseline methods. The tensorrt+taesd+VFB setup achieves 10.05 FPS while consuming only 104-110W, compared to 205W for the vanilla baseline—representing a 47\% reduction in power consumption while delivering 5.4× speedup, translating to approximately 11× better performance per watt~\cite{schwartz2019greenai}. Even Xformer-based configurations demonstrate favorable power profiles, with xformer+taesd+VFB consuming 153W at 8.0 FPS versus the baseline's 196W at 2.22 FPS.
Memory efficiency constitutes another critical advantage of StreamFlow. Our optimized pipelines reduce peak memory usage from 3788MB (vanilla baseline) to 2666-3012MB across various configurations, achieving 20-30\% memory savings through our batched velocity field processing strategy, which eliminates redundant intermediate tensor allocations. The TensorRT configurations achieve the lowest memory footprint (2666-2694MB) due to aggressive operator fusion and memory planning~\cite{chen2018tvmautomatedendtoendoptimizing,dao2022flashattentionfastmemoryefficientexact}. Combined with throughput improvements, our framework enables serving 5-6× more requests per GPU, dramatically improving infrastructure utilization and cost efficiency in production environments.

\subsection{Scalability Assessment}

To evaluate the robustness of our acceleration framework under varying computational demands, we conduct a scalability study across different image resolutions ranging from 64px to 1024px. As illustrated in Figure \ref{fig:scable}, our method demonstrates remarkable stability compared to baseline approaches, maintaining substantial speedup ratios even as resolution increases. Under TensorRT compilation (Fig. \ref{fig:scable:a}, \ref{fig:scable:b}), our tensorrt+taesd+VFB configuration achieves consistent 4-5× acceleration across all resolutions in both text-to-image and img2img modes. The Xformer-based configurations (Fig. \ref{fig:scable:c}, \ref{fig:scable:d}) exhibit even stronger scalability—at 1024px resolution, while baseline methods experience severe performance collapse, our xformer+taesd+VFB configuration maintains nearly 80\% of its peak throughput. This resilience stems from our decoupled velocity field processing, which adapts naturally to increased tensor dimensions without introducing architectural bottlenecks, validating that StreamFlow provides genuine algorithmic-level acceleration rather than exploiting resolution-specific optimizations.

\begin{table}[t]
\centering
\caption{\textbf{Quality Study:} Higher ClIP score is better; lower FID are better. It can be seen that our acceleration scheme has almost no impact on the generation quality (the impact is less than 1\%).}
\label{tab:quality_for_gen}
\resizebox{\linewidth}{!}{%
\begin{tabular}{lccc}
\toprule
Method & ClIP score $\uparrow$ & FID $\downarrow$ & VAE \\
\midrule
tensorrt+taesd+VFB       & 96.25 & 31.28 & taesd \\
tensorrt+Original VAE+VFB           & 96.08 & 31.16 & Original VAE \\
xformer+taesd+VFB   &  97.26 & 30.75 & taesd \\
xformer+Original VAE+VFB            & 96.74 &  31.82 & Original VAE \\
\midrule
Baseline HF~\cite{von-platen-etal-2022-diffusers}              &  97.27 & 30.08 & Original VAE \\
Baseline Vanilla~\cite{yan2024perflow}         &  97.28 & 29.44 & Original VAE \\
\bottomrule
\end{tabular}
}
\vspace{-12pt}
\end{table}

\subsection{Quality Assessment}

To verify that our acceleration framework does not compromise generation quality, we evaluate CLIP Score and FID on standard text-to-image benchmarks. As shown in Table 2, our method maintains generation quality comparable to the baselines, with less than 1\% variation in CLIP Score. 
These results confirm that StreamFlow's core acceleration mechanisms preserve generation fidelity while delivering substantial speedups, making our framework suitable for production deployment where both speed and quality are critical.
\section{Conclusion}

In this paper, we present the first comprehensive acceleration framework specifically designed for accelerating the Rectified Flow model. We have thoroughly analyzed the difficulties (Velocity Field Batching, Heterogeneous Timesteps, Dynamic Compilation) in designing a new acceleration framework for flow models from a theoretical, strategic, and practical perspective. We have also understood the reasons for the failure of traditional methods and proposed corresponding solutions. Our acceleration framework is plug-and-play and can significantly increase the generation speed of 512-dimensional images of the Rectified Flow model by 611\% with almost no loss in quality. 
Even on larger dimensions, we can still achieve an almost 4-5 times improvement, and in this case, the native Hugging Face pipeline is almost unable to accelerate. Compared with the inconvenient-to-reuse methods such as distillation, Reflow, and few-step, we directly make the deployment of large-scale flow models possible.

\section*{Impact Statement}

This paper presents work whose goal is to advance the field of Machine
Learning. There are many potential societal consequences of our work, none
which we feel must be specifically highlighted here.

\nocite{langley00}

\bibliography{ref/main}
\bibliographystyle{icml2026}



\end{document}